\documentclass[draftclasofoot, onecolumn]{IEEEtran}
\IEEEoverridecommandlockouts

\usepackage[noadjust]{cite}
\usepackage[bookmarks=true,breaklinks=true,colorlinks,linkcolor=red,citecolor=blue,urlcolor=BlueViolet]{hyperref}
\usepackage{multirow}
\usepackage{amsmath,amssymb,amsfonts}
\usepackage[ruled,vlined]{algorithm2e}
\newcommand{\mycommentstyle}[1]{\color[HTML]{0671b9}{#1}}
\SetKwComment{Comment}{\mycommentstyle{// }}{}
\usepackage{graphicx}
\usepackage{threeparttable}
\usepackage{subfigure}
\usepackage{textcomp}
\usepackage{xcolor}
\def\BibTeX{{\rm B\kern-.05em{\sc i\kern-.025em b}\kern-.08em
    T\kern-.1667em\lower.7ex\hbox{E}\kern-.125emX}}
\usepackage{booktabs}
\usepackage{tablefootnote}
\usepackage{blindtext}
\usepackage{float}
\usepackage{algorithmic}
\usepackage{textcomp}
\usepackage{booktabs}
\usepackage{tabularx}
\usepackage{tensor}

\begin{document}

\title{Neural Tucker Convolutional Network for Water Quality Analysis}

\author{\IEEEauthorblockN{Hongnan Si}\\
	\IEEEauthorblockA{\textit{College of Computer and Information Science}
		\textit{Southwest University}
		Chongqing, China \\
		171212337@qq.com}
\\
	\IEEEauthorblockN{Tong Li}\\
	\IEEEauthorblockA{\textit{College of Software}
		\textit{Handan University}
		Hebei, China \\
		chestnut1xxx@gmail.com}
\\
	\IEEEauthorblockN{Yujie Chen}\\
	\IEEEauthorblockA{\textit{Rural Revitalization Service Center of Hechuan}
		Chongqing, China \\
		584987025@qq.com}
\\
	\IEEEauthorblockN{Xin Liao*}\\
	\IEEEauthorblockA{\textit{College of Computer and Information Science}
		\textit{Southwest University}
		Chongqing, China \\
		lxchat@email.swu.edu.cn}
}

\maketitle

\begin{abstract}
Water quality monitoring is a core component of ecological environmental protection. However, due to sensor failure or other inevitable factors, data missing often exists in long-term monitoring, posing great challenges in water quality analysis. This paper proposes a Neural Tucker Convolutional Network (NTCN) model for water quality data imputation, which features the following key components: a) Encode different mode entities into respective embedding vectors, and construct a Tucker interaction tensor by outer product operations to capture the complex mode-wise feature interactions; b) Use 3D convolution to extract fine-grained spatiotemporal features from the interaction tensor. Experiments on three real-world water quality datasets show that the proposed NTCN model outperforms several state-of-the-art imputation models in terms of accuracy.
\end{abstract}


\section{Introduction}
In advancing the modernization drive for harmonious coexistence between humans and nature, water quality monitoring plays an irreplaceable role \cite{r1,r2,r3,r4,r5,r6, b1}. Therefore, accurately collecting water quality-related data is particularly important for making scientific and reasonable work arrangements—it can identify pollution sources, assess ecological risks, and formulate effective response measures \cite{r7,r8,r9,r10,r11,r12, b2}. However, practical data collection faces many challenges \cite{r13,r14,r15,r16,r17,r18, b3}. As the scope of monitoring expands and the number of devices increases, data gaps have emerged. These missing pieces of information prevent us from fully understanding the real state of water quality \cite{r19,r20,r21,r22,r23,r24, b4}, so how to accurately fill in these missing data has become an important issue that urgently needs to be addressed \cite{r25,r26,r27,r28,r29,r30, b5}.

In recent years, researchers have proposed various methods to address the missing value imputation problem of high-dimensional and sparse data. For example, traditional tensor decomposition methods such as Canonical Polyadic (CP) and Tucker Decomposition (TD)~\cite{r31,r32,r33,r34,r35,r36, b6} can only capture the multilinear correlations of data and fail to characterize nonlinear interaction relationships. In neural networks models, NTF can only model the time dimension and is difficult to capture local correlations in the spatial dimension (e.g., the correlation between water quality monitoring stations and water quality indicators)~\cite{r37,r38,r39,r40,r41,r42, b7}. Decision trees are employed to model the complex relationships among features and capture nonlinear correlations, thereby enabling accurate data imputation. Nevertheless, this method is prone to the risk of overfitting~\cite{r43,r44,r45, b8}.

Therefore, we propose the Neural Tucker Convolutional Network (NTCN) model, which combines embedding representation, 3D convolutional operations, and a multi-layer perceptron (MLP) network. It can effectively capture complex nonlinear relationships, thereby improving imputation accuracy. The main contributions of this paper are as follows:
\begin{enumerate}
	\item 3D convolutional feature extraction mechanism. By performing two-layer 3D convolutional operations on the high-order interaction tensor constructed via outer product of embeddings for monitoring stations, water quality indicators, and time slices, it effectively captures the high-order interaction features inherent in the tensor, laying a foundation for accurate imputation.
	\item Embedding regularization. The embedding weights of monitoring stations, water quality indicators, and time slices are initialized with a uniform distribution, which helps the model converge faster and avoid local optima. Meanwhile, combined with L2 regularization, it reduces the risk of overfitting.
	\item Comparative experiments based on multiple datasets. Experimental results show that the proposed model outperforms traditional imputation methods in terms of the evaluation metrics RMSE and MSE.
\end{enumerate}

The organizational structure of the remaining sections is as follows: Section 
\ref{sec:pre} introduces the notation definitions and problem formulation; Section \ref{sec:method} elaborates on the detailed design of the proposed NTCN model; Section \ref{sec:empirical} describes the experimental setup and the results of comparative experiments; Section \ref{sec:conclusion} summarizes the full paper and outlines future research directions.

\begin{figure*}[!ht]
	\centering
	\includegraphics[width=1\textwidth]{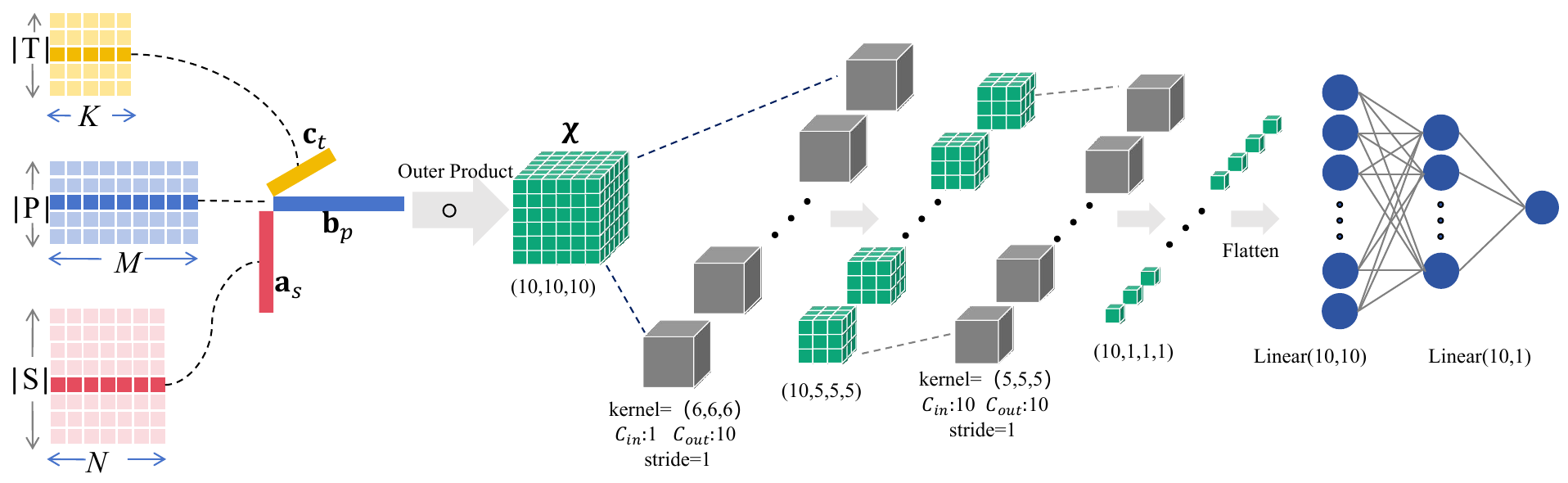}
	\caption{Architecture of the NTCN model for water quality data imputation. The numerical labels in the figure correspond to the parameter configurations employed in the experiments, including kernel size, stride, and number of channels.}
	\label{fig:1} 
	\vspace{-5pt} 
\end{figure*}

\section{Preliminary}
\label{sec:pre}
\subsection{Notation system}
The notations employed in this paper are summarized in Table \uppercase\expandafter{\romannumeral1}.
\begin{table}[htbp]
	\centering
	\scriptsize
	\caption{Symbol Description}
	\begin{tabularx}{\linewidth}{l X}
		\toprule 
		\textbf{Symbol} & \textbf{Description} \\
		\midrule 
		$\mathrm{S}$ & Set of monitoring stations, $\mathrm{S} \! = \!\{s_1, s_2, ..., s_N\}$ \\
		$\mathrm{P}$ & Set of water quality indicators (e.g., pH, dissolved oxygen, etc.), $\mathrm{P} = \{p_1, p_2, ..., p_M\}$ \\
		$\mathrm{T}$ & Set of time slices, $\mathrm{T} =\{t_1, t_2, ..., t_K\}$ \\
		$\boldsymbol{\mathcal{Y}} \in \mathbb{R}^{\mathrm{|S|} \times \mathrm{|P|} \times \mathrm{|T|}}$ & Raw water quality tensor, ${y}_{spt}$ represents the observation on water quality indicator $p$ at station $s$ in time slice $t$ \\
		$\hat{\boldsymbol{\mathcal{Y}}} \in \mathbb{R}^{\mathrm{|S|} \times \mathrm{|P|} \times \mathrm{|T|}}$ & Water quality data tensor after imputation \\
		$\mathbf{a}_s \in \mathbb{R}^N$ & Embedding vector of monitoring station $s$ with dimension $N$ \\
		$\mathbf{b}_p \in \mathbb{R}^M$ & Embedding vector of water quality indicator $p$ with dimension $M$ \\
		$\mathbf{c}_t \in \mathbb{R}^K$ & Embedding vector of time slice $t$ with dimension $K$ \\
		$\boldsymbol{\mathcal{W}}_1 \in \mathbb{R}^{c_1 \times 1 \times k_1 \times k_1 \times k_1}$ & Weight of the first 3D convolution layer, $c_1$ is the number of output channels,
		$k_1$ is the kernel size \\
		$\boldsymbol{\mathcal{W}}_2 \in \mathbb{R}^{c_2 \times c_1 \times k_2 \times k_2 \times k_2}$ & Weight of the second 3D convolution layer, $c_2$ is the number of output channels, $k_2$ is the kernel size \\
		$\mathbf{W}_3 \in \mathbb{R}^{h_1 \times c_{in}}$ & Weight of the first layer of the Multilayer Perceptron (MLP), $c_{in}$ is the input dimension, $h_1$ is the hidden layer dimension \\
		$\mathbf{W}_4 \in \mathbb{R}^{1 \times h_1}$ & Weight of the second layer of the MLP \\
		$|\cdot|$ & The cardinality of a set \\
		\bottomrule 
	\end{tabularx}
	\label{tab:symbol_description}
\end{table}

\subsection{Problem formulation}
The WQD can be represented by an HDS tensor,which is defined as:

\textbf{Definition 1}. (HDS tensor). For a given tensor $\boldsymbol{\mathcal{Y}} \in \mathbb{R}^{|\mathrm{S}|\times|\mathrm{P}|\times|\mathrm{T}|}$, let $\Omega$ and $\Upsilon$ denote the observed and unobserved element sets in the tensor, respectively. If $|\Upsilon| \gg |\Omega|$, the tensor $\boldsymbol{\mathcal{Y}}$ is an HDS tensor~\cite{r46,r47,r48,r49,r50, b9}.

\textbf{Definition 2}. (Tucker Decomposition). TD decomposes into a core tensor multiplied by a factor matrix along each mode, which is formulated as:
\begin{equation}
	\label{eq:td}
	\boldsymbol{\mathcal{Y}} \approx [\![\boldsymbol{\mathcal{G}}; \mathbf{A}, \mathbf{B}, \mathbf{C}]\!] = \boldsymbol{\mathcal{G}} \times_1 \mathbf{A} \times_2 \mathbf{B} \times_3 \mathbf{C},
\end{equation}
where $\mathbf{A} \in \mathbb{R}^{|\mathrm{S}|\times N}$, $\mathbf{B} \in \mathbb{R}^{|\mathrm{P}|\times M}$ and $\mathbf{C} \in \mathbb{R}^{|\mathrm{T}|\times K}$ are latent factor matrices with ranks of $N$, $M$ and $K$, and elements of core tensor $\boldsymbol{\mathcal{G}}$ show the level of interactions among different latent factors. By the definition, the elementwise form of the TD is

\begin{equation}
	\label{eq:tdd}
	{y}_{spt} \approx \hat{y}_{spt} = \sum_{n=1}^{N} \sum_{m=1}^{M} \sum_{k=1}^{K} 
	g_{nmk} a_{sn} b_{pm} c_{tk}.
\end{equation}

\section{The NTCN model}
\label{sec:method}
The overall architecture of the proposed NTCN model is illustrated in Fig. \ref{fig:1}, which consists of three core modules: embedding-based high-order interaction tensor construction fusing monitoring stations, water quality indicators and time slices features via outer product, 3D convolutional feature extraction, and MLP prediction.

\subsection{Embedding and Tensor Construction}
For the water quality tensor $\boldsymbol{\mathcal{Y}}$, each element is mapped to the interval $[0, 1]$ using the Sigmoid function.
\begin{equation}
	\label{eq:sgf}
	y'_{spt} = \sigma(y_{spt}) = \frac{1}{1+e^{-{y}_{spt}}}.
\end{equation}

Through embedding layers, the IDs of monitoring stations, water quality indicators, and time slices corresponding to observed values are mapped into the dense latent feature vectors, thereby addressing the feature representation issue of HDS data:

\begin{equation}
	\label{eq:wjo}
	\begin{cases}
		\mathbf{a}_s = \mathcal{E}(s), \\
		\mathbf{b}_p = \mathcal{E}(p), \\
		\mathbf{c}_t = \mathcal{E}(t),
	\end{cases}
\end{equation}
where $\mathbf{a}_s \in \mathbb{R}^{N}$, $\mathbf{b}_p \in \mathbb{R}^{M}$, $\mathbf{c}_t \in \mathbb{R}^{K}$ and $\mathcal{E}(\cdot)$ denotes the embedding process.

The outer product operation~\cite{r51,r52,r53,r54,r55, b10} is used to fuse the embedding vectors of the monitoring stations, water quality indicators, and time slices, constructing a three-order feature tensor to capture the high-order interactions among the three mode~\cite{r56,r57,r58,r59,r60,b11, b14}:
\begin{equation}
	\label{eq:wjf}
	\boldsymbol{\mathcal{X}}_{spt} = \mathbf{a}_s \circ \mathbf{b}_p \circ \mathbf{c}_t,
\end{equation}
where $\circ$ denotes the outer product operation, and $\boldsymbol{\mathcal{X}}_{spt} \in \mathbb{R}^{N \times M \times K}$ of which each element is defined as $x^{(nmk)}_{spt} = a_{sn} \cdot b_{pm} \cdot c_{tk}$. 

To enhance the robustness on sparse data, the tensor is standardized as:
\begin{equation}
	\label{eq:wjstandard}
	\boldsymbol{\mathcal{X}}_{spt}' = \frac{\boldsymbol{\mathcal{X}}_{spt} - \mu_{{\mathcal{X}}}}{\sigma_{{\mathcal{X}}} + \epsilon},
\end{equation}
where $\mu_{\mathcal{X}}$ and $\sigma_{\mathcal{X}}$ are the mean and standard deviation of the tensor, respectively, and $\epsilon = 1 \times 10^{-8}$ is a small constant to prevent division by zero.

\subsection{3D Convolutional Feature Extraction}

High-order spatiotemporal interaction features are extracted from the tensor through two layers of 3D convolution:
\begin{equation}
	\label{eq:cvo}
	\boldsymbol{\mathcal{H}}_1 = \text{ReLU}\left( \text{Conv3D}( \boldsymbol{\mathcal{X}}_{spt}^{'}, \boldsymbol{\mathcal{W}}_1)\right) ,
\end{equation}
\begin{equation}
	\label{eq:cvt}
	\boldsymbol{\mathcal{H}}_2 = \text{ReLU}\left( \text{Conv3D}(\boldsymbol{\mathcal{H}}_1, \boldsymbol{\mathcal{W}}_2) \right) ,
\end{equation}
where $\boldsymbol{\mathcal{H}}_1 \in \mathbb{R}^{c_1 \times d_{11} \times d_{12} \times d_{13}}$ and $\boldsymbol{\mathcal{H}}_2 \in \mathbb{R}^{c_2 \times d_{21} \times d_{22} \times d_{23}}$ is the convolutional feature tensor extracted via the respective convolution layer. In this paper, the two 3D convolution kernel have size of $k_1=6$ and $k_2=5$. The $\operatorname{ReLU}(\cdot)$ denotes the ReLU activation function with the form of $\operatorname{ReLU}(x) = \max(0, x)$.

\subsection{MLP Prediction and Regularization}

The convolutional feature tensor $\boldsymbol{\mathcal{H}}_2$ is flattened and then mapped into the predicted value of the water quality indicator via a two-layer MLP structure~\cite{r61,r62,r63,r64,r65, b12, b13}:
\begin{equation}
	\label{eq:fl}
	\mathbf{h}_2 = \operatorname{Flatten}(\boldsymbol{\mathcal{H}}_2) \in \mathbb{R}^n,
\end{equation}
\begin{equation}
	\label{eq:rlt}
	\mathbf{h}_3 = \operatorname{ReLU}(\mathbf{W}_3 \mathbf{h}_2 + \mathbf{b}_3) \in \mathbb{R}^m,
\end{equation}
\begin{equation}
	\label{eq:yc}
	\hat{y}_{spt} = \sigma(\mathbf{W}_4 \mathbf{h}_3 + b_o) \in (0, 1),
\end{equation}
where $\sigma(\cdot)$ is the sigmoid function which ensures the output lies within the $[0, 1]$, matching with the range of preprocessed water quality data.

The objective function of this model consists of two parts: the mean square error loss (MSE) and L2 regularization. Its specific form is as follow:
\begin{equation}
	\label{eq:aim}
	\begin{split}
		\mathcal{L} &= \mathcal{L}_{o} + \lambda \cdot \mathcal{L}_{reg}(\Theta)\\
		&=\frac{1}{2} \sum_{y_{spt} \in \Omega} (y_{spt} - \hat{y}_{spt})^2 + \lambda \cdot \Theta^2,
	\end{split}     
\end{equation}

where $\lambda$ is the regularization coefficient, and $\Theta$ represents all learnable parameters.

The specific process of the model is presented in Algorithm 1.

\begin{algorithm}[htbp]
	\caption{NTCN Model}
	\label{alg:conv_tucker_simple}
	\SetAlgoLined
	\KwIn{$\boldsymbol{\mathcal{Y}}$, $c_1, c_2$, $k_1, k_2$, $\lambda$, $T$}
	\KwOut{$\hat{\boldsymbol{\mathcal{Y}}}$}
	Initialize $\mathbf{a}_s, \mathbf{b}_p, \mathbf{c}_t \sim U(0, 0.004)$, $\boldsymbol{\mathcal{W}}_1, \boldsymbol{\mathcal{W}}_2, \mathbf{W}_3, \mathbf{W}_4$\;
	$t = 1$\;
	\While{$t \leq T$}{
		\For{each quadruple $(s, p, t,y_{spt}) \in \Omega$}{
			$\boldsymbol{\mathcal{X}}_{spt} = \mathbf{a}_s \circ \mathbf{b}_p \circ \mathbf{c}_t$, \quad
			
			$\boldsymbol{\mathcal{X}} \leftarrow$ standardize $\boldsymbol{\mathcal{X}}$ according to \eqref{eq:wjstandard};
			
			$\boldsymbol{\mathcal{H}_1} = \operatorname{Conv3D}(\boldsymbol{\mathcal{X}'}, \boldsymbol{\mathcal{W}}_1)$, 
			
			$\boldsymbol{\mathcal{H}_2} = \operatorname{Conv3D}(\boldsymbol{\mathcal{H}_1},  \boldsymbol{\mathcal{W}_2})$, 
			
			$\hat{y}_{spt} = \operatorname{MLP}(\boldsymbol{\mathcal{H}_2})$\;
			Backpropagation according to \eqref{eq:aim};
		}
		$t = t + 1$\;
	}
	\Return{$\hat{\boldsymbol{{\mathcal{Y}}}}$}
\end{algorithm}

\section{Experimental Comparisons}
\label{sec:empirical}
\subsection{Datasets}\label{AA}
Based on three sets of real-world water quality data from 24 monitoring stations in Victoria Harbour established by the Environmental Protection Department of Hong Kong: surface layer D1 with 129,406 records, middle layer D2 with 121,218 records, and bottom layer D3 with 129,415 records. This study covers 24 water quality-related parameters over 903 time slices. The datasets are divided into training set ($\Gamma$), validation set ($\Lambda$), and test set ($\Phi$) sets at a ratio of 1:2:7 to evaluate the performance of the proposed model.

\subsection{Evaluation metrics}
To measure the model's performance, this study employs the common Root Mean Square Error (RMSE) and Mean Absolute Error (MAE) as evaluation metrics~\cite{r66,r67,r68,r69,r70,r71,b15,b16}:
\begin{equation}
	\label{eq:mae}
	\text{MAE} = \frac{1}{|\Phi|} \sum_{y_{spt} \in \Phi}  |y_{spt} - \hat{y}_{spt} | ,
\end{equation}
\begin{equation}
	\label{eq:rmse}
	\text{RMSE} = \sqrt{\frac{1}{|\Phi|} \sum_{{y_{spt} \in \Phi}} \left(y_{spt} - \hat{y}_{spt} \right)^2}.
\end{equation}

\subsection{Comparison Results}
We compare the results of the proposed NTCN model with four state-of-the-art
imputation models on three datasets in terms of imputation accuracy~\cite{r72,r73,r74,r75,r76, b17, b18}. The tested modoels are: a) M1: The proposed NTCN model; b) M2: A biased low-rank tensor model based on non-negative multiplicative updates; c) M3: An imputation model that employs the Cauchy loss function and handles outliers without explicit outlier detection; d) M4: A multi-dimensional prediction model based on tensor decomposition and the reconfiguration optimization algorithm; e) M5: A model adopting a tensor completion mechanism that combines CP decomposition, standard gradient descent and alternating least squares optimization.

\begin{table}[htbp]
	\centering
	\caption{Imputation Performance Comparison}
	\label{tab: imputation_accuracy}
	\begin{tabular}{c c c c c c c}
		\toprule
		Dataset & & M1 & M2 & M3 & M4 & M5 \\
		\midrule
		\multirow{2}{*}{D1} & RMSE & {0.0243} & 0.0256 & 0.0260 & 0.0273 & 0.0267 \\
		& MAE  & {0.0149} & 0.0162 & 0.0148 & 0.0162 & 0.0163 \\
		\multirow{2}{*}{D2} & RMSE & {0.0259} & 0.0262 & 0.0266 & 0.0287 & 0.0270 \\
		& MAE  & {0.0168} & 0.0175 & 0.0161 & 0.0177 & 0.0176 \\
		\multirow{2}{*}{D3} & RMSE & {0.0249} & 0.0258 & 0.0262 & 0.0277 & 0.0259 \\
		& MAE  & {0.0164} & 0.0176 & 0.0159 & 0.0175 & 0.0166 \\
		\bottomrule
	\end{tabular}
\end{table}

To evaluate its performance, the proposed model is benchmarked  against four current best-performing models across three datasets, with a focus on imputation accuracy.
\

The rank of all models are set to 10 and the iteration rounds are set to 1000. Note that the tested models are terminated if the error in two consecutive iteration rounds is less than $10^{-5}$. Tables \uppercase\expandafter{\romannumeral2} present the statistical results, and it can be observed that: 
\begin{enumerate}
	\item The NTCN model significantly outperforms other tested models in terms of imputation accuracy. As shown in Table II, on the D1 dataset, the NTCN model M1 achieves an RMSE of 0.0243 and an MSE of 0.0149. Compared with M2, which has an RMSE of 0.0256 and MSE of 0.0162; M4, which has an RMSE of 0.0273 and an MSE of 0.0162; and M5, which has an RMSE of 0.0267 and MSE of 0.0163, the MSE of M1 is reduced by 8.02\%, 8.02\%, and 8.59\% respectively. It only differs from M3's MSE of 0.0148 by 0.0001, being basically the same and maintaining the optimal level.
	\item On the D2 dataset, the RMSE of M1 is significantly lower than that of other models, and the MAE of M1 is 0.0168. Compared with M2's MAE of 0.0175, M4's MAE of 0.0177, and M5's MAE of 0.0176, it is reduced by 4.00\%, 5.08\%, and 4.55\% respectively. It is only slightly higher than M3's 0.0161, and still in the leading echelon overall.
	\item On the D3 dataset, the MAE of M1 is 0.0164. Compared with M2's MAE of 0.0176, M4's MAE of 0.0175, and M5's MAE of 0.0166, it is reduced by 6.82\%, 6.29\%, and 1.20\% respectively. It only differs from M3's 0.0159 by 0.0005, maintaining advantages as well. In terms of RMSE, M1 is the lowest among the three datasets, fully reflecting the absolute advantage of the model in imputation accuracy.
\end{enumerate}

In summary, the proposed NTCN model achieves high accuracy in imputing missing values of water quality data. It can accurately capture the high-order interaction relationships among monitoring stations, water quality indicators, and time slices, making both RMSE and MAE better than those of most comparison models, thus providing an efficient and reliable solution for practical water quality monitoring data processing.

\section{Conclusions}
\label{sec:conclusion}
To achieve accurate imputation of missing data, this paper proposes a 3D convolution-based model. Specifically, the model maps the features of monitoring stations, water quality indicators, and time slices to a low-dimensional latent space through embedding layers, and constructs a 3D feature tensor using outer product operations to capture the high-order interaction relationships among the three. On this basis, the model adopts two-layer 3D convolutional operations for feature extraction and dimensionality reduction of the tensor. Leveraging the local perception capability of convolutional kernels, it mines local correlation patterns in the latent space, and then completes nonlinear mapping and final prediction through a MLP. The advantages of the model are as follows: on the one hand, it reduces the scale of model parameters while retaining key interaction information, effectively alleviating the overfitting risk caused by HDS data; on the other hand, the 3D convolutional operation can capture the complex correlations of features among the monitoring station, water quality indicator, and time slice. Performance evaluation on water quality datasets shows that the proposed model outperforms existing imputation models in missing data imputation tasks~\cite{r77,r78,r79,r80,b19,b20}. In the future, we plan to explore on air quality monitoring datasets and verify its application potential in broader ecological sensing tasks.

\bibliographystyle{IEEEtran}
\bibliography{CLR}

@article{r1,
  title={Neural Nonnegative Latent Factorization of Tensors Model With Acceleration and Unconstraint},
  author={Li, Wenqiang and Lin, Mingwei and Xu, Xiuqin and Lin, Ling and Xu, Zeshui and Luo, Xin},
  journal={IEEE Transactions on Systems, Man, and Cybernetics: Systems},
  year={2025},
  publisher={IEEE}
}

@article{r2,
  title={An Intelligent Optimization-Based Residual Negative Magnitude Shaping Scheme for Vibration Control},
  author={Yang, Weiyi and Li, Shuai and Luo, Xin},
  journal={IEEE Transactions on Industrial Electronics},
  year={2025},
  publisher={IEEE}
}

@article{r3,
  title={Dynamic Stochastic Reorientation Particle Swarm Optimization for Adaptive Latent Factor Analysis in High-Dimensional Sparse Matrices},
  author={Lyu, Chao and Ma, Ziwen and Luo, Xin and Shi, Yuhui},
  journal={IEEE Transactions on Knowledge and Data Engineering},
  year={2025},
  publisher={IEEE}
}

@article{r4,
  title={Learning Accurate Representation to Nonstandard Tensors via a Mode-Aware Tucker Network},
  author={Wu, Hao and Wang, Qu and Luo, Xin and Wang, Zidong},
  journal={IEEE Transactions on Knowledge and Data Engineering},
  year={2025},
  publisher={IEEE}
}

@article{r5,
  title={A Convolution Bias-Incorporated Nonnegative Latent Factorization of Tensors Model for Accurate Representation Learning to Dynamic Directed Graphs},
  author={Wang, Qu and Wu, Hao and Luo, Xin},
  journal={IEEE Transactions on Systems, Man, and Cybernetics: Systems},
  year={2025},
  publisher={IEEE}
}

@article{r6,
  title={A Proximal-ADMM-Incorporated Nonnegative Latent-Factorization-of-Tensors Model for Representing Dynamic Cryptocurrency Transaction Network},
  author={Liao, Xin and Wu, Hao and He, Tiantian and Luo, Xin},
  journal={IEEE Transactions on Systems, Man, and Cybernetics: Systems},
  year={2025},
  publisher={IEEE}
}

@article{r7,
  title={Discovering Spatiotemporal--Individual Coupled Features From Nonstandard Tensors—A Novel Dynamic Graph Mixer Approach},
  author={Bi, Fanghui and He, Tiantian and Ong, Yew-Soon and Luo, Xin},
  journal={IEEE Transactions on Neural Networks and Learning Systems},
  year={2025},
  publisher={IEEE}
}

@article{r8,
  title={A novel tensor causal convolution network model for highly-accurate representation to spatio-temporal data},
  author={Liao, Xin and Wu, Hao and Luo, Xin},
  journal={IEEE Transactions on Automation Science and Engineering},
  year={2025},
  publisher={IEEE}
}

@article{r9,
  title={A Fine-Grained Regularization Scheme for Nonnegative Latent Factorization of High-Dimensional and Incomplete Tensors},
  author={Wu, Hao and Qiao, Yan and Luo, Xin},
  journal={IEEE Trans. Serv. Comput.},
 volume={17},
  number={6},
  pages={3006--3021},
  year={2024},
  publisher={IEEE}
}

@article{r10,
  title={Temporal pattern-aware QoS prediction by biased non-negative Tucker factorization of tensors},
  author={Tang, Peng and Ruan, Tao and Wu, Hao and Luo, Xin},
  journal={Neurocomputing},
  volume={582},
  pages={127447},
  year={2024},
  publisher={Elsevier}
}

@article{r11,
  title={Advancing non-negative latent factorization of tensors with diversified regularization schemes},
  author={Wu, Hao and Luo, Xin and Zhou, MengChu},
  journal={IEEE Transactions on Services Computing},
  volume={15},
  number={3},
  pages={1334--1344},
  year={2020},
  publisher={IEEE}
}

@article{r12,
  title={A PID-incorporated latent factorization of tensors approach to dynamically weighted directed network analysis},
  author={Wu, Hao and Luo, Xin and Zhou, MengChu and Rawa, Muhyaddin J and Sedraoui, Khaled and Albeshri, Aiiad},
  journal={IEEE/CAA Journal of Automatica Sinica},
  volume={9},
  number={3},
  pages={533--546},
  year={2021},
  publisher={IEEE}
}

@article{r13,
  title={Temporal pattern-aware QoS prediction via biased non-negative latent factorization of tensors},
  author={Luo, Xin and Wu, Hao and Yuan, Huaqiang and Zhou, MengChu},
  journal={IEEE transactions on cybernetics},
  volume={50},
  number={5},
  pages={1798--1809},
  year={2019},
  publisher={IEEE}
}

@article{r14,
  title={A novel approach to large-scale dynamically weighted directed network representation},
  author={Luo, Xin and Wu, Hao and Wang, Zhi and Wang, Jianjun and Meng, Deyu},
  journal={IEEE Transactions on Pattern Analysis and Machine Intelligence},
  volume={44},
  number={12},
  pages={9756--9773},
  year={2021},
  publisher={IEEE}
}

@article{r15,
  title={NeuLFT: A novel approach to nonlinear canonical polyadic decomposition on high-dimensional incomplete tensors},
  author={Luo, Xin and Wu, Hao and Li, Zechao},
  journal={IEEE Transactions on Knowledge and Data Engineering},
  year={2022},
  publisher={IEEE}
}

@inproceedings{r16,
  title={Neural latent factorization of tensors for dynamically weighted directed networks analysis},
  author={Wu, Hao and Luo, Xin and Zhou, MengChu},
  booktitle={2021 IEEE International Conference on Systems, Man, and Cybernetics (SMC)},
  pages={3061--3066},
  year={2021},
  organization={IEEE}
}

@inproceedings{r17,
  title={Instance-frequency-weighted regularized, nonnegative and adaptive latent factorization of tensors for dynamic QoS analysis},
  author={Wu, Hao and Luo, Xin},
  booktitle={2021 IEEE International Conference on Web Services (ICWS)},
  pages={560--568},
  year={2021},
  organization={IEEE}
}

@inproceedings{r18,
  title={Discovering hidden pattern in large-scale dynamically weighted directed network via latent factorization of tensors},
  author={Wu, Hao and Luo, Xin and Zhou, MengChu},
  booktitle={2021 IEEE 17th International Conference on Automation Science and Engineering (CASE)},
  pages={1533--1538},
  year={2021},
}

@inproceedings{r19,
  title={Proportional-integral-derivative-incorporated latent factorization of tensors for large-scale dynamic network analysis},
  author={Wu, Hao and Xia, Yan and Luo, Xin},
  booktitle={2021 China Automation Congress (CAC)},
  pages={2980--2984},
  year={2021},
  organization={IEEE}
}

@inproceedings{r20,
  title={Efficient representation to dynamic QoS data via generalized nesterov’s accelerated gradient-incorporated biased non-negative latent factorization of tensors},
  author={Chen, Minzhi and Luo, Xin},
  booktitle={2021 IEEE International Conference on Systems, Man, and Cybernetics (SMC)},
  pages={576--581},
  year={2021},
  organization={IEEE}
}

@article{r21,
  title={Adjusting learning depth in nonnegative latent factorization of tensors for accurately modeling temporal patterns in dynamic QoS data},
  author={Luo, Xin and Chen, Minzhi and Wu, Hao and Liu, Zhigang and Yuan, Huaqiang and Zhou, MengChu},
  journal={IEEE Transactions on Automation Science and Engineering},
  volume={18},
  number={4},
  pages={2142--2155},
  year={2021},
  publisher={IEEE}
}

@article{r22,
  author={Chen, Hong and Lin, Mingwei and Zhao, Liang and Xu, Zeshui and Luo, Xin},
  journal={IEEE Transactions on Intelligent Transportation Systems}, 
  title={Fourth-Order Dimension Preserved Tensor Completion With Temporal Constraint for Missing Traffic Data Imputation}, 
  year={2025},
  volume={26},
  number={5},
  pages={6734-6748},
}

@book{r23,
  title={Dynamic Network Representation Based on Latent Factorization of Tensors},
  author={Wu, Hao and Wu, Xuke and Luo, Xin},
  year={2023},
  publisher={Springer}
}

@article{b1,
  title={A particle swarm with local decision algorithm for functional distributed constraint optimization problems},
  author={Shi, Meifeng and Liao, Xin and Chen, Yuan},
  journal={International journal of pattern recognition and artificial intelligence},
  volume={36},
  number={12},
  pages={2259025},
  year={2022},
  publisher={World Scientific}
}

@article{r24,
  title={Local Search-based Anytime Algorithms for Continuous Distributed Constraint Optimization Problems},
  author={Liao, Xin and Hoang, Khoi and Luo, Xin},
  journal={IEEE/CAA Journal of Automatica Sinica},
  volume={12},
  number={1},
  pages={288--290},
  year={2025},
  publisher={IEEE}
}

@article{b2,
  title={A dual-population search differential evolution algorithm for functional distributed constraint optimization problems},
  author={Shi, Meifeng and Liao, Xin and Chen, Yuan},
  journal={Annals of Mathematics and Artificial Intelligence},
  volume={90},
  number={10},
  pages={1055--1078},
  year={2022},
  publisher={Springer}
}

@inproceedings{r25,
  title={Sgd-dyg: Self-reliant global dependency apprehending on dynamic graphs},
  author={Han, Minglian and Wang, Ling and Yuan, Ye and Luo, Xin},
  booktitle={Proceedings of the 31st ACM SIGKDD Conference on Knowledge Discovery and Data Mining V. 2},
  pages={802--813},
  year={2025}
}

@article{b3,
  title={Biased Block Term Tensor Decomposition for Temporal Pattern-aware QoS Prediction},
  author={Wang, Qu and Liao, Xin and Wu, Hao},
  journal={International Journal of Pattern Recognition and Artificial Intelligence},
  year={2025},
  publisher={World Scientific}
}

@inproceedings{b4,
  title={An Adaptive Temporal-Dependent Tensor Low-Rank Representation Model for Dynamic Communication Network Embedding},
  author={Liao, Xin and Hu, Qicong and Tang, Peng},
  booktitle={2024 International Conference on Networking, Sensing and Control (ICNSC)},
  pages={1--6},
  year={2024},
  organization={IEEE}
}

@ARTICLE{r26,
  author={Lin, Mingwei and Liu, Jiaqi and Chen, Hong and Xu, Xiuqin and Luo, Xin and Xu, Zeshui},
  journal={IEEE Transactions on Intelligent Transportation Systems}, 
  title={A 3D Convolution-Incorporated Dimension Preserved Decomposition Model for Traffic Data Prediction}, 
  year={2025},
  volume={26},
  number={1},
  pages={673-690},
}

@article{r27,
  title={An Adaptive Neighborhood-Resonated Graph Convolution Network for Undirected Weighted Graph Representation},
  author={Chen, Jiufang and Yuan, Ye and Luo, Xin and Gao, Xinbo},
  journal={IEEE Transactions on Neural Networks and Learning Systems},
  year={2025},
  publisher={IEEE}
}

@article{r28,
  title={Auto-encoding neural tucker factorization},
  author={Tang, Peng and Luo, Xin and Woodcock, Jim},
  journal={IEEE Transactions on Knowledge and Data Engineering},
  year={2025},
  publisher={IEEE}
}

@article{r29,
  title={Neural Networks-Incorporated Latent Factor Analysis for High-Dimensional and Incomplete Data},
  author={Lin, Mingwei and Lin, Xingyu and Xu, Xiuqin and Xu, Zeshui and Luo, Xin},
  journal={IEEE Transactions on Systems, Man, and Cybernetics: Systems},
  year={2025},
  publisher={IEEE}
}

@article{r30,
  title={Identifying novel therapeutic targets of natural compounds in traditional Chinese medicine herbs with hypergraph representation learning},
  author={Qiao, Yantong and Hu, Lun and Zhang, Jun and Hu, Pengwei and Luo, Xin},
  journal={Briefings in Bioinformatics},
  volume={26},
  number={4},
  pages={bbaf399},
  year={2025},
  publisher={Oxford University Press}
}

@article{r31,
  title={FMvPCI: A Multiview Fusion Neural Network for Identifying Protein Complex via Fuzzy Clustering},
  author={Yang, Yue and Hu, Lun and Li, Guodong and Li, Dongxu and Hu, Pengwei and Luo, Xin},
  journal={IEEE Transactions on Systems, Man, and Cybernetics: Systems},
  year={2025},
  publisher={IEEE}
}

@article{r32,
  title={MNL: A highly-efficient model for large-scale dynamic weighted directed network representation},
  author={Chen, Minzhi and He, Chunlin and Luo, Xin},
  journal={IEEE Transactions on Big Data},
  year={2022},
  publisher={IEEE}
}

@article{r33,
  title={Low-rank high-order tensor completion with applications in visual data},
  author={Qin, Wenjin and Wang, Hailin and Zhang, Feng and Wang, Jianjun and Luo, Xin and Huang, Tingwen},
  journal={IEEE Transactions on Image Processing},
  volume={31},
  pages={2433--2448},
  year={2022},
  publisher={IEEE}
}

@article{r34,
  title={A proportional integral controller-enhanced non-negative latent factor analysis model},
  author={Yuan, Ye and Lu, Siyang and Luo, Xin},
  journal={IEEE/CAA Journal of Automatica Sinica},
  volume={12},
  number={6},
  pages={1246--1259},
  year={2025},
  publisher={IEEE}
}

@article{r35,
  title={Link-Based Attributed Graph Clustering via Approximate Generative Bayesian Learning},
  author={Yang, Yue and Hu, Lun and Li, Guodong and Li, Dongxu and Hu, Pengwei and Luo, Xin},
  journal={IEEE Transactions on Systems, Man, and Cybernetics: Systems},
  year={2025},
  publisher={IEEE}
}

@article{r36,
  title={Enhancing graph convolutional networks with an efficient k-hop neighborhood approach},
  author={Chen, Jiufang and Luo, Xin and Yuan, Ye and Wang, Zidong},
  journal={Information Fusion},
  pages={103297},
  year={2025},
  publisher={Elsevier}
}

@article{r37,
  title={From data analysis to intelligent maintenance: a survey on visual defect detection in aero-engines},
  author={Wu, Peishu and Li, Han and Luo, Xin and Hu, Liwei and Yang, Rui and Zeng, Nianyin},
  journal={Measurement Science and Technology},
  volume={36},
  number={6},
  pages={062001},
  year={2025},
  publisher={IOP Publishing}
}

@article{r38,
   author={Xu, Xiuqin and Lin, Mingwei and Luo, Xin and Xu, Zeshui},
  journal={IEEE Transactions on Services Computing}, 
  title={An Adaptively Bias-Extended Non-Negative Latent Factorization of Tensors Model for Accurately Representing the Dynamic QoS Data}, 
  year={2025},
  volume={18},
  number={2},
  pages={603-617},
}

@article{r39,
  title={Nonnegative Latent Factor Analysis-Incorporated and Feature-Weighted Fuzzy Double $ c $-Means Clustering for Incomplete Data},
  author={Song, Yan and Li, Ming and Zhu, Zhengyu and Yang, Guisong and Luo, Xin},
  journal={IEEE Transactions on Fuzzy Systems},
  volume={30},
  number={10},
  pages={4165--4176},
  year={2022},
  publisher={IEEE}
}

@article{r40,
  title={Robust Low-Rank Latent Feature Analysis for Spatiotemporal Signal Recovery},
  author={Wu, Di and Li, Zechao and Yu, Zhikai and He, Yi and Luo, Xin},
  journal={IEEE Transactions on Neural Networks and Learning Systems},
  year={2023},
  publisher={IEEE}
}

@article{r41,
  title={Proximal alternating-direction-method-of-multipliers-incorporated nonnegative latent factor analysis},
  author={Bi, Fanghui and Luo, Xin and Shen, Bo and Dong, Hongli and Wang, Zidong},
  journal={IEEE/CAA Journal of Automatica Sinica},
  volume={10},
  number={6},
  pages={1388--1406},
  year={2023},
  publisher={IEEE}
}

@article{r42,
  title={A Kalman-filter-incorporated latent factor analysis model for temporally dynamic sparse data},
  author={Yuan, Ye and Luo, Xin and Shang, Mingsheng and Wang, Zidong},
  journal={IEEE Transactions on Cybernetics},
  year={2022},
  publisher={IEEE}
}

@article{r43,
  title={Pseudo Gradient-Adjusted Particle Swarm Optimization for Accurate Adaptive Latent Factor Analysis},
  author={Luo, Xin and Chen, Jiufang and Yuan, Ye and Wang, Zidong},
  journal={IEEE Transactions on Systems, Man, and Cybernetics: Systems},
  year={2024},
  publisher={IEEE}
}

@article{r44,
  title={A Fast Nonnegative Autoencoder-based Approach to Latent Feature Analysis on High-Dimensional and Incomplete Data},
  author={Bi, Fanghui and He, Tiantian and Luo, Xin},
  journal={IEEE Transactions on Services Computing},
  year={2023},
  publisher={IEEE}
}

@article{r45,
  title={Attention-Mechanism-Based Neural Latent-Factorization-of-Tensors Model},
  author={Xu, Xiuqin and Lin, Mingwei and Xu, Zeshui and Luo, Xin},
  journal={ACM Transactions on Knowledge Discovery from Data},
  volume={19},
  number={4},
  pages={1--27},
  year={2025},
}

@article{r46,
  title={Position-transitional particle swarm optimization-incorporated latent factor analysis},
  author={Luo, Xin and Yuan, Ye and Chen, Sili and Zeng, Nianyin and Wang, Zidong},
  journal={IEEE Transactions on Knowledge and Data Engineering},
  volume={34},
  number={8},
  pages={3958--3970},
  year={2020},
  publisher={IEEE}
}

@article{r47,
  title={A double-space and double-norm ensembled latent factor model for highly accurate web service QoS prediction},
  author={Wu, Di and Zhang, Peng and He, Yi and Luo, Xin},
  journal={IEEE Transactions on Services Computing},
  volume={16},
  number={2},
  pages={802--814},
  year={2022},
  publisher={IEEE}
}

@article{r48,
  title={A prediction-sampling-based multilayer-structured latent factor model for accurate representation to high-dimensional and sparse data},
  author={Wu, Di and Luo, Xin and He, Yi and Zhou, Mengchu},
  journal={IEEE Transactions on Neural Networks and Learning Systems},
  volume={35},
  number={3},
  pages={3845--3858},
  year={2022},
  publisher={IEEE}
}

@article{r49,
  title={An algorithm of inductively identifying clusters from attributed graphs},
  author={Hu, Lun and Yang, Shicheng and Luo, Xin and Zhou, MengChu},
  journal={IEEE Transactions on Big Data},
  volume={8},
  number={2},
  pages={523--534},
  year={2020},
  publisher={IEEE}
}

@article{r50,
 title={Momentum-accelerated and Biased Unconstrained Non-negative Latent Factor Model for Handling High-Dimensional and Incomplete Data},
  author={Lin, Mingwei and Yang, Hengshuo and Xu, Xiuqin and Lin, Ling and Xu, Zeshui and Luo, Xin},
  journal={ACM Transactions on Knowledge Discovery from Data},
  year={2025},
  publisher={ACM New York, NY}
}

@article{r51,
  title={An alternating-direction-method of multipliers-incorporated approach to symmetric non-negative latent factor analysis},
  author={Luo, Xin and Zhong, Yurong and Wang, Zidong and Li, Maozhen},
  journal={IEEE Transactions on Neural Networks and Learning Systems},
  volume={34},
  number={8},
  pages={4826--4840},
  year={2021},
  publisher={IEEE}
}

@article{r52,
  title={A momentum-accelerated Hessian-vector-based latent factor analysis model},
  author={Li, Weiling and Luo, Xin and Yuan, Huaqiang and Zhou, MengChu},
  journal={IEEE Transactions on Services Computing},
  volume={16},
  number={2},
  pages={830--844},
  year={2022},
  publisher={IEEE}
}

@article{r53,
  title={Two-stream graph convolutional network-incorporated latent feature analysis},
  author={Bi, Fanghui and He, Tiantian and Xie, Yuetong and Luo, Xin},
  journal={IEEE Transactions on Services Computing},
  volume={16},
  number={4},
  pages={3027--3042},
  year={2023},
  publisher={IEEE}
}

@article{r54,
  title={A Fuzzy PID-Incorporated Stochastic Gradient Descent Algorithm for Fast and Accurate Latent Factor Analysis},
  author={Yuan, Ye and Li, Jinli and Luo, Xin},
  journal={IEEE Transactions on Fuzzy Systems},
  year={2024},
  publisher={IEEE}
}

@article{r55,
  title={Alternating-Direction-Method of Multipliers-Based Adaptive Nonnegative Latent Factor Analysis},
  author={Zhong, Yurong and Liu, Kechen and Gao, Shangce and Luo, Xin},
  journal={IEEE Transactions on Emerging Topics in Computational Intelligence},
  year={2024},
  publisher={IEEE}
}

@article{r56,
  title={SDGNN: Symmetry-Preserving Dual-Stream Graph Neural Networks},
  author={Chen, Jiufang and Yuan, Ye and Luo, Xin},
  journal={IEEE/CAA Journal of Automatica Sinica},
  volume={11},
  number={7},
  pages={1717--1719},
  year={2024},
  publisher={IEEE}
}

@article{r57,
  title={A distributed adaptive second-order latent factor analysis model},
  author={Wang, Jialiang and Li, Weiling and Luo, Xin},
  journal={IEEE/CAA Journal of Automatica Sinica},
  year={2024},
  publisher={IEEE}
}

@article{r58,
  title={Asynchronous Parallel Fuzzy Stochastic Gradient Descent for High-Dimensional Incomplete Data Representation},
  author={Qin, Wen and Luo, Xin},
  journal={IEEE Transactions on Fuzzy Systems},
  year={2023},
  publisher={IEEE}
}

@article{r59,
  title={Adaptively-accelerated Parallel Stochastic Gradient Descent for High-Dimensional and Incomplete Data Representation Learning},
  author={Qin, Wen and Luo, Xin and Zhou, MengChu},
  journal={IEEE Transactions on Big Data},
  year={2023},
  publisher={IEEE}
}

@article{r60,
  title={Saliency-aware dual embedded attention network for multivariate time-series forecasting in information technology operations},
  author={Li, Jiajia and Tan, Feng and He, Cheng and Wang, Zikai and Song, Haitao and Hu, Pengwei and Luo, Xin},
  journal={IEEE Transactions on Industrial Informatics},
  year={2023},
  publisher={IEEE}
}

@article{r61,
  title={Fast and accurate non-negative latent factor analysis of high-dimensional and sparse matrices in recommender systems},
  author={Luo, Xin and Zhou, Yue and Liu, Zhigang and Zhou, MengChu},
  journal={IEEE Transactions on Knowledge and Data Engineering},
  volume={35},
  number={4},
  pages={3897--3911},
  year={2021},
  publisher={IEEE}
}

@article{r62,
  title={Predicting protein-protein interactions using sequence and network information via variational graph autoencoder},
  author={Luo, Xin and Wang, Liwei and Hu, Pengwei and Hu, Lun},
  journal={IEEE/ACM Transactions on Computational Biology and Bioinformatics},
  volume={20},
  number={5},
  pages={3182--3194},
  year={2023},
  publisher={IEEE}
}

@article{r63,
  title={FCAN-MOPSO: an improved fuzzy-based graph clustering algorithm for complex networks with multiobjective particle swarm optimization},
  author={Hu, Lun and Yang, Yue and Tang, Zehai and He, Yizhou and Luo, Xin},
  journal={IEEE Transactions on Fuzzy Systems},
  volume={31},
  number={10},
  pages={3470--3484},
  year={2023},
  publisher={IEEE}
}

@article{r64,
  title={A graph-incorporated latent factor analysis model for high-dimensional and sparse data},
  author={Wu, Di and He, Yi and Luo, Xin},
  journal={IEEE Transactions on Emerging Topics in Computing},
  year={2023},
  publisher={IEEE}
}

@article{r65,
  title={Highly accurate manipulator calibration via extended Kalman filter-incorporated residual neural network},
  author={Yang, Weiyi and Li, Shuai and Li, Zhibin and Luo, Xin},
  journal={IEEE Transactions on Industrial Informatics},
  volume={19},
  number={11},
  pages={10831--10841},
  year={2023},
  publisher={IEEE}
}

@article{r66,
  title={Diversified regularization enhanced training for effective manipulator calibration},
  author={Li, Zhibin and Li, Shuai and Bamasag, Omaimah Omar and Alhothali, Areej and Luo, Xin},
  journal={IEEE Transactions on Neural Networks and Learning Systems},
  volume={34},
  number={11},
  pages={8778--8790},
  year={2022},
  publisher={IEEE}
}

\vspace{12pt}

\end{document}